\documentclass[sigconf]{acmart}

\renewcommand\footnotetextcopyrightpermission[1]{} % removes footnote with conference information in first column
\usepackage{booktabs} % For formal tables
\usepackage{amsthm}
\usepackage{algorithm}
\usepackage[noend]{algorithmic}
\usepackage{adjustbox}

\newcommand{\stitle}[1]{\vspace{1ex} \noindent{\bf #1}}

\begin{document}
\title{Joint Embedding in Named Entity Linking on Sentence Level}
%\titlenote{Produces the permission block, and
%  copyright information}
%\subtitle{Extended Abstract}
%\subtitlenote{The full version of the author's guide is available as
%  \texttt{acmart.pdf} document}

\author{Wei Shi}
\affiliation{
    \institution{The Chinese University of Hong Kong}}
\email{shiw@se.cuhk.edu.hk}

\author{Siyuan Zhang}
\affiliation{
    \institution{The Chinese University of Hong Kong}}
\email{syzhang@se.cuhk.edu.hk}

\author{Zhiwei Zhang}
\affiliation{
    \institution{Hong Kong Baptist University}}
\email{cszwzhang@comp.hkbu.edu.hk}

\author{Hong Cheng}
\affiliation{
    \institution{The Chinese University of Hong Kong}}
\email{hcheng@se.cuhk.edu.hk}

\author{Jeffrey Xu Yu}
\affiliation{
    \institution{The Chinese University of Hong Kong}}
\email{yu@se.cuhk.edu.hk}

\long\def\comment#1{}

\begin{abstract}
Named entity linking is to map an ambiguous mention in documents
to an entity in a knowledge base.
The named entity linking is challenging, given the fact that there are
multiple candidate entities for a mention in a document.  It is
difficult to link a mention when it appears multiple times in a
document, since there are conflicts by the contexts around the
appearances of the mention.  In addition, it is
difficult since the given training dataset is small due to the reason
that it is done manually to link a mention to its mapping entity.
In the literature, there are many reported studies among which the
recent embedding methods learn vectors of entities from the training
dataset at document level.
To address these issues, we focus on how to link entity for mentions
at a sentence level, which reduces the noises introduced by different
appearances of the same mention in a document at the expense of
insufficient information to be used. We propose a new unified
embedding method by maximizing the relationships learned from
knowledge graphs. We confirm the effectiveness of our method
in our experimental studies.
\end{abstract}


\maketitle

\section{Introduction}

Entity linking is an important issue in understanding of ambiguous
texts given knowledge bases such as DBpedia
\cite{DBLP:journals/ws/BizerLKABCH09} and YAGO
\cite{DBLP:conf/www/SuchanekKW07} induced by Wikipedia.  Consider a
Wikipedia article, where there are phrases marked by underline, called
mentions, for instance, ``Cambridge'', which may link to an entity of
``University of Cambridge'', or link to an entity of ``Cambridge,
Massachusetts'', depending on where the word of ``Cambridge''
appears. Such a pair of mention and entity in a Wikipedia is called an
anchor. Such a mention (e.g., ``Cambridge'') is ambiguous since it
refers to a different thing in a different context.
The process of mapping an ambiguous mention to the correct entity in a
knowledge base is the task of named entity linking.

Entity linking is first proposed in Wikify
\cite{DBLP:conf/cikm/MihalceaC07}. Wikify proposes two algorithms. One
is inspired by Lesk \cite{DBLP:conf/sigdoc/Lesk86}
to compute the overlap of words between an entity description in
Wikipedia articles and the paragraph of a mention.
The other uses a Naive Bayes classifier using context features.
To improve the accuracy of named entity linking, the existing methods
focus on three things, which are prior probability of candidate
entity, context similarity between mentions and candidate entities, and
coherence among candidate entities of different mentions. For prior
probability,
the statistical method, that computes the prior probability based on
the anchor pairs in Wikipedia articles, is the most frequently
selected method
\cite{DBLP:journals/tacl/LingSW15,DBLP:conf/conll/YamadaS0T16,DBLP:conf/emnlp/HoffartYBFPSTTW11,DBLP:conf/i-semantics/MendesJGB11,DBLP:conf/acl/HuHDGX15}.
For context similarity,
\cite{DBLP:conf/emnlp/HoffartYBFPSTTW11} uses keyphrase-based
similarity and syntax-based similarity.
\cite{DBLP:conf/conll/YamadaS0T16} uses the word embedding of all the
noun words in a document as the context feature and puts it into
Gradient Boosted Regression Trees (GBRT)
\cite{friedman2001greedy}. DBpedia Spotlight
\cite{DBLP:conf/i-semantics/MendesJGB11} applies a vector space model
to measure the similarity between the context of mention and candidate
entity.
For coherence
\cite{DBLP:journals/tacl/LingSW15,
DBLP:conf/naacl/PershinaHG15,
DBLP:conf/www/JehW03,
DBLP:conf/emnlp/HoffartYBFPSTTW11,
DBLP:conf/conll/YamadaS0T16},
Graph-based
disambiguation algorithms are largely adopted
for improving the impact of coherence.
\cite{DBLP:conf/emnlp/HoffartYBFPSTTW11} designs a
mention-entity graph and formalizes coherence as a Steiner-tree
problem and then gives a greedy algorithm which maximizes the minimum
degree of the vertices.
\cite{DBLP:conf/conll/YamadaS0T16} calculates the similarity between
the vector representation of candidate entities of ambiguous mentions
and the vector representation of unambiguous mentions.

Following the recently development of embedding methods on knowledge graphs,
there are works that apply embedding algorithms on entity
linking \cite{DBLP:conf/acl/HuHDGX15,DBLP:conf/conll/YamadaS0T16}.
However, they do not take contexts of mentions,
relations among entities, and coherence information into consideration
together.
In \cite{DBLP:conf/conll/YamadaS0T16}, it uses the skip-gram model to
jointly learn entity-entity, word-word, and entity-word
embeddings. However, all the three embeddings are taken from the
Wikipedia articles and are only combined by the word and entity in
anchors in Wikipedia articles.

The named entity linking is challenging for several reasons, given the
fact that there are multiple candidate entities for a mention in a
document. First,
%
% ({\bf Issue-1})
%
it is difficult to link a mention when it appears multiple times in a
document, since there are conflicts by the contexts around the
appearances of the mention.
Second, the given training dataset is small, since it is done manually
to link a mention to its mapping entity.
In addition, not all mapping entities in the test data can be found in the training dataset.
To address the first, in this work, we link entity for mentions at a
sentence level, which reduces the noises introduced by different
appearances of the same mention in a document at the expense of
insufficient information to be used.
To address the second and third, we utilize a large number of anchors (e.g.,
mention-entity pairs) extracted from the Wikipedia articles to
augment the training dataset and use knowledge graphs to capture various relationships among entities.
To summarize, we propose a novel method to jointly embed all the
three components, which are contexts of mentions,
relations among entities and coherence information, together into a high-dimensional space to solve the named entity linking task at sentence level. Our
experimental studies confirm the effectiveness of our approach.

\begin{figure*}[t]
  \includegraphics[width=\textwidth,height=4.6cm]{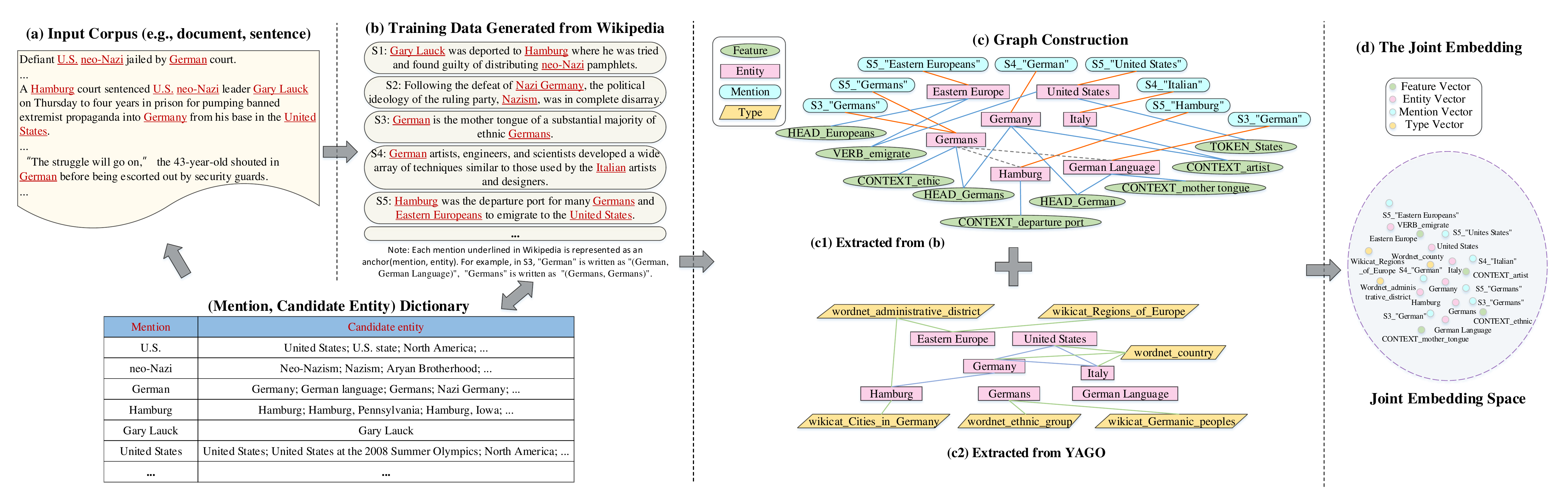}
\caption{The Framework}
\vspace*{-0.2cm}
\label{elfig:ekframe}
\end{figure*}

\vspace*{-0.2cm}
\section{The Problem} \label{elsec.pff}

Consider a collection of documents $D$. For each document $d$ in $D$,
a set of mentions, denoted as $M(d)$, are marked (or labeled).  A
mention $m$ is modeled by context features, denoted as $F_C(m)$. There
are 10 context features:
``head of the mention'', ``tokens'', ``context unigrams'', ``context
bigrams'', ``part-of-speech tags'', ``word shape'', ``length'',
``character sequence'' and ``brown clusters'', as given in
\cite{DBLP:conf/aaai/LingW12} and ``nearest verb''. We explain it using 
Example~\ref{exa.fs}. 
In the following, we use $F$ to represent all the context features for all mentions used in model learning.

\begin{example} \label{exa.fs}
Consider the sentence S3 in Figure~\ref{elfig:ekframe}: ``German is the mother tongue of a substantial majority of ethnic Germans.''
Here ``German'' is a mention. Its context features are shown
in Table~\ref{eltab.mentionfeatures}.
\end{example}

\begin{table*}[t]
%\small
\footnotesize
  \caption{Context Features of Mentions}
  \vspace*{-0.2cm}
  \label{eltab.mentionfeatures}
  \adjustbox{max width=\textwidth}{
  \centering
  \begin{tabular}{lll}
    \toprule
    \textbf{Feature} & \textbf{Description} & \textbf{Example} \\
    \midrule
    Head of the mention & The head of the mention following the rules by \cite{DBLP:conf/emnlp/CollinsS99} & ``HEAD\_German'' \\
    Tokens & The words (sometimes stopwords are dropped) in the mention & ``German'' \\
    Context unigrams & The tokens in a context window of the mention & ``mother'', ``tongue'' \\
    Context bigrams & The context bigrams including the mention & `` German mother'', ``mother tongue'' \\
    Part-of-Speech tags & The part-of-speech tags of the mention & ``NN'' \\
    Word Shape & The word shape of the tokens in the mention & ``Aa'' for ``German'' \\
    Length & The length of the mention & 1 \\
    Character sequence & The continuous character sequences in the mention & ``Ger'', ``erm'', ``rma'', ``man'' \\
    Brown clusters & The cluster id of each token in the mention (using the first 8-bit prefixes) & ``8\_11101100'' \\
    Nearest verb & The nearest verb to the mention & ``be'' \\
    \bottomrule
  \end{tabular}}
\end{table*}

The problem is to find the entity in a knowledge base for a mention in
a document.
Here, a knowledge graph (e.g., YAGO) models knowledge using
``(subject, predicate, object)'' tuples where a subject/object can be an
entity or category (also called type), and a predicate is to specify the
the relationship between a subject and an object.
For example, (Boston, is-a, City), states that Boston is a city, where
Boston is an entity and City is a category.
The knowledge graph can be modeled as a labeled graph $G = (V, E)$,
where $V$ is a set of vertices to represent entities and categories,
and $E$ is a set of edges between two vertices. Edge labels are used
to specify what the relationships (predicates) are.

Every mention, $m_i$ in $M(d)$, corresponds to a set of candidate
entities denoted as $E(m_i) = \{e_{i1}, e_{i2}, \cdots\}$, where
$e_{ij}$ is a vertex in the knowledge graph $G$.
The named entity linking problem is to identify one entity $e_{ij}$
from $E(m_i)$ in the knowledge graph $G$ for every mention $m_i$ in
every document $d$ in $D$, given a training set of documents $T$ in
which the entity for every mention in a document in $T$ is identified.
For an entity $e_{ij}$ that correctly links to a mention $m_i$, we
call the entity $e_{ij}$ a mapping entity for $m_i$ below.

The problem is challenging for several reasons, given the fact that
there are multiple candidate entities for a mention in a
document.
({\bf Issue-1}) A mention may appear several times in different
sentences in a document, and an appearance of a mention may link to a
different entity.  For example, in Figure~\ref{elfig:ekframe}(a),
there are two appearance of ``German'' in a document that map to
different entities. The first appearance maps to ``Germany'' and the
second appearance maps to ``German Language''.  If we take all
information from a document as a whole to identify the mapping entity
for a multi-appearance mention, it becomes extremely difficult to
identify the mapping entity for each appearance of a mention. If we
only take information from a sentence where a mention appears, the
information obtained may be too small to link it to the mapping
entity.
({\bf Issue-2}) The training dataset $T$ is small, since it is done
manually to link a mention to its mapping entity.
({\bf Issue-3}) Not all mapping entities of mentions in the dataset of $D$ to be tested
can be found in the training dataset $T$. In some datasets, only a
half of mapping entities in $D$ can be found in $T$.

%%%%%%%%%%%%%%%%%%%%%%%%%%%%%

The state-of-art method is \cite{DBLP:conf/conll/YamadaS0T16} which
uses three kinds of features, base features, string similarity
features, and contextual features.
The base features include four features, namely, the prior probability
$p(e|m)$ of a candidate entity $e$ given a mention $m$, the entity
prior $p(e)$, the maximum prior probability of the candidate entity of
all mentions in a document, and the number of entity candidates for a
mention. Among them, the prior probability is the most prominent
in named entity linking.
The string similarity features include three features, namely, the
edit distance between a candidate entity and its mention, whether the
candidate entity is identical to or contains its mention, whether the
candidate entity starts or ends with its mention.
The contextual features, that are computed based on the learned
embeddings of words and entities, contain three features, the cosine
similarity between a candidate entity and the textual context of its
mention, the cosine similarity between a candidate entity and
contextual entities, and the descending order of a candidate entity
among all candidate entities of its mention according to the sum of
these two contextual similarities.
To get the embeddings of words and entities, it makes use of the large
amounts of articles in Wikipedia. It embeds (1) word-word occurrence
relations, (2) entity-entity inlink relations and (3) the word-entity
co-occurrence relations, by the skip-gram model, respectively.  Here,
for (2), it is done by taking inlink entities as context words and the
linked entity as the target word. For (3), recall that an anchor in a
Wikipedia is a mention-entity pair,
the words are those that appear in such mentions of anchors.
Together, it learns the embeddings of words and entities jointly.
Given the embedding learned, it further uses
Gradient Boosted Regression Trees (GBRT) \cite{friedman2001greedy},
which is a prediction model in the form of an ensemble of regression
trees.
It achieves high performance using {\tt aida-yago2-dataset}.  However,
when a mention appears multiple times in a document, it only links to
one entity.
As an example, as shown in Figure~\ref{elfig:ekframe}(a), the first
appearance of ``German'' should be linked to ``Germany'', where the
second appearance of ``German'' should be linked to ``German
language''. By \cite{DBLP:conf/conll/YamadaS0T16}, it links to either
``Germany'' or ``German language'', but not both.  This is because it
uses all the words and entities that appear in the entire document
together as its contextual information at a document level.

\vspace*{-0.2cm}
\section{Our Approach}
\label{sec:ours}

Figure~\ref{elfig:ekframe} shows our framework. There are 3 main
steps.
First, for a given input corpus, $D$, we obtain mention-entity pairs
from $D$ using a dictionary available, which keeps a list of candidate
entities for a mention. We denote the dictionary for $D$ as $Dict$ in
which an entry keeps a list of candidate entities for a mention in
$D$. Then we extract Wikipedia articles that can be used as training
dataset. Recall that there are anchors in a Wekipedia article, which
is a (mention, entity) pair. An Wikipedia article is selected if it
contains mentions in $Dict$ (or in $D$) above a given threshold.
We select those sentences from a Wikipedia article selected if
they contain at least a mention in $Dict$.
Second, we construct a graph as shown in
Figure~\ref{elfig:ekframe}(c). We represent the graph in 2 parts.  In
Figure~\ref{elfig:ekframe}(c1), it shows the graph constructed that
represents (i) the relationships between context features and
entities, (ii) the relationships between mentions and entities, and
(iii) the co-occurrence relationship between two entities if the
corresponding mentions appear in the same sentence (denoted as dashed
lines).  Here, the context features are extracted around the anchors
in the training sentences as shown in Figure~\ref{elfig:ekframe}(b).
In Figure~\ref{elfig:ekframe}(c2), it shows the graph constructed that
represents the relationships (i) between entities and (ii) between
entities and types using a knowledge graph (e.g., YAGO).  In the
following we denote the graph constructed as $G_{D}$ (for
the input corpus $D$).
Third, we learn a model by a joint embedding model
(see Figure~\ref{elfig:ekframe}(d)) based on $G_{D}$.
With the embedding learned, we link a mention to the mapping entity by
comparing the similarity between the vector representation of the
mention and the vector representations of the candidate entities.

To address Issue-1, we do entity linking at a sentence level. On one
hand, it reduces the noise introduced by different appearances of the
same mention in a document. On the other hand, the information that
can be used is reduced.
To represent a mention, we extract various context features around a
mention to capture the mention, since it is most likely that, for the
same mention, the context features of two appearances are different.
By treating mentions by their context features, we aim at linking the
same mention to different entities if they appear in different
contexts.
For example, as shown in Figure~\ref{elfig:ekframe} (b), for
``German'' in the sentence S3, there is a feature ``mother tongue'',
while for ``German'' in the sentence S4, there is a feature
``artist''.
By the relations obtained by embedding between the feature of ``mother
tongue'' and the entity ``German Language'', as shown in
Figure~\ref{elfig:ekframe}(d), the mention ``German'' in S3 is closer
to the entity ``German Language'' than ``Germany'' in the embedding
space.
To address Issue-2 and Issue-3, we utilize a large numbers of anchors
contained in the Wikipedia articles, where an anchor is considered as
a mention-entity pair, and in addition we capture the relationships
among entity by a knowledge graph $G$.  Here, Wikipedia provides large
quantities of manually labeled anchors (e.g., mention-entity
pairs). Such high-quality mention-entity pairs can be used as training
dataset.
The knowledge graph contains
high-quality semantic relations between entities and between entities
and types (e.g., the ``is-a'' relation between entities and types).

Below, we discuss (1) how to get a large training dataset, $D_L$, and
(2) how to learn a model from $D_L$.

We construct a dictionary, $Dict$, from the page titles, the disambiguation and redirect
pages of Wikipedia, in which the candidate entities for every
mention is maintained as $Dict = \{(m, V_{m})\}$, where $m$ is a mention
and $V_{m}$ is the set of candidate entities for $m$. For a given training dataset $T$,
we use $M$ to represent all mentions in $T$ as $M = \cup_{d \in T}
M(d)$, where $M(d)$ is the set of mentions in a document $d$. We use $Dict$ to find the candidate entities for $\forall m \in M$ and obtain the set of mention-candidate pairs for $T$ which is given as $D_{MCE} =
\{(m, e) | m \in M \land e \in V_{m} \land (m, V_{m}) \in Dict\}$.

The set of mention-candidate pairs, $D_{MCE}$, is rather small to test
all mentions in testing.
We generate a large training data $D_{L}$ from an external source
(e.g. Wikipedia) based on $T$. The Wikipedia is huge and
contains unnecessary information for the domain to be tested.
First, we consider an article, $d'$, in Wikipedia as a candidate to be
selected. It is worth mentioning that there are many the so-called
anchors in a Wikipedia article that can be considered as a pair of
(mention, entity), where the mention in an anchor is the words appear
in the article and the entity is the one to be linked to. The
connections among such entities can be determined from the knowledge
graph $G$ behind (e.g., YOGA2).
A Wikipedia article containing a mention $m$, that appears in $T$,
will be selected as $d'$, if the number of inlinks and outlinks is
greather than a given threshold. We then treat every anchor in $d'$ as
a (mention, entity) pair.
Let $T'$ be the set of such documents $\{d'\}$, which is much larger
than a given training dataset $T$. Hence, we can get a large set of
mention-candidate pairs from $T'$ and a large $D'_{MCE} = \{(m',
e')\}$, if the anchor (or the mention-entity pair) of $(m', e')$
appears in a document $d'$ in $T'$.

Next, we discuss our approach to learn a model from a training dataset
$D_L$ by an unified embedding which consists of 4 embeddings, namely,
(1) feature-entity embedding to capture the co-occurrence relations
between features and entities, (2) mention-entity embedding to capture
the correct mappings, (3) knowledge graph embedding to capture the
semantic relations between entities as well as ``is-a'' relation between
entities and types, and (4) the mention-entity
embedding to capture the coherence relations in context.
Recall the training dataset $D_L$ is a set of tuples where a tuple in
$D_L$ is for a mention in a sentence such as $(s, m, mid, e)$.  Here,
$s$ is a sentence, $m$ is a mention identified by $mid$, and $e$ is
the corresponding entity to be linked. It is important to note that
any mention in a sentence may link to an entity which is irrelevant to
the same mention if it appears in a different sentence. In other
words, consider $(s, m, mid, e)$ and $(s', m', mid', e')$, assuming $s
\neq s'$ and $m = m'$. We treat $m$ and $m'$ differently and explore
if they link to different entities, $e$ and $e'$, by identifying them
differently (i.e., $mid \neq mid'$).

\stitle{Feature-Entity Embedding}: Since a mention, $m$, in a
sentence, $s$, may link to an entity $e$, independent on the
appearance of $m$ in other sentences, we represent a mention by its
context feature (e.g., words), $f$, in the sentence where the mention
$m$ appears. %An entity $e$ may be linked to $m$ if the similarity between $m$ and $e$ is high.
A mention $m$ may be linked to $e$ if the similarity between $f$ and $e$ is high.
Let $F_C(m)$ be the context features (e.g., the words) around $m$. %, and
%$F_C(e)$ be the context features around $e$.
The similarity between
$m$ and $e$ are measured by the similarity between $F_C(m)$ and $e$. %$F_C(e)$.
It is important to note that the context features serve as a
bridge between a mention and an entity.
We use the skip-gram model of Word2vec, which shows the highly
relatedness between two words or phrases if they share many of the
same context words, to carry the co-occurrence relations between
features and entities, in the sense (1) the more entities two
features share the more similar they tend to be and (2)
the more features two entities share the more similar they tend.
As a result, if two mentions share more common features or similar
features, it is more likely that they map to the similar entities.
Based on the discussion made above, we discuss feature-entity
embedding below.
Consider a vector representation in a $d$-dimensional embedding space
for a feature $f$ in the set of entire features $F$, and an enity $e$
that appears in a tuple $(s, m, mid, e)$ in the training dataset
$D_{L}$. More precisely, let ${\textbf{f}}$ and ${\textbf{y}}$ be
vectors for $f$ and $e$ respectively.
The co-occurrence relations between features and entities are shown in
Eq.~(\ref{elequ.fecoh}) by the skip-gram model.
\begin{equation} \label{elequ.fecoh}
\setlength{\abovedisplayskip}{3pt}
\setlength{\belowdisplayskip}{3pt}
FE = - \sum_{f \in F} \sum_{e \in D_{L}} w \cdot \log p(e | f)
\end{equation}
where $p(e | f)$ is the probability of $e$ generated by $f$ such as
$p(e | f) = exp(\textbf{f}^{T}\textbf{y}) / \sum_{e' \in D_{L}}
exp(\textbf{f}^{T}\textbf{y}')$, and $w$ is the co-occurrence
frequency between $f$ and $e$ in $D_{L}$.

To achieve efficiency,
in this wrok, we use negative sampling
\cite{DBLP:conf/nips/MikolovSCCD13} to sample various false features
for each $(f, e)$ based on the widely used noise distribution
$P_{n}(e) \propto D_{e}^{\frac{3}{4}}$ where $D_{e}$ indicates the
number of co-occurrence time between features and $e$
\cite{DBLP:conf/nips/MikolovSCCD13}.
By such sampling, the probability $p(e | f)$ in
Eq.~(\ref{elequ.fecoh}) can be computed by Eq.~(\ref{elequ.fens}).
\vspace*{-0.3cm}
\begin{equation} \label{elequ.fens}
\setlength{\abovedisplayskip}{3pt}
\setlength{\belowdisplayskip}{3pt}
p(e | f) = \log \sigma(\textbf{f}^{T}\textbf{y}) + \sum_{k =
  1}^{Q}\Xi_{e'\thicksim P_{n}(e)}[\log
  \sigma(-\textbf{f}^{T}\textbf{y}')]
\end{equation}
where $\sigma(x) = 1 / (1 + exp(-x))$ is the sigmoid function, $\Xi$
is a distribution, and $Q$ is the number of negative samples.

\stitle{Mention-Entity Embedding}: In general, the same mention is
possibly linked to different entities if it appears in different
contexts. In a tuple of $(s, m, mid, e)$ in $D_{L}$, we use $mid$ to
unique identify a mention $m$ in the sentence $s$.
To abuse the notation, in the following we use $mid$ to indicate the
mention identified by $mid$. In other words, $mid$ is considered as a
unique mention instead of an identifier.
For example, let $\phi$ be a similarity function, we use $\phi(mid,
e)$ to indicate the similarity between $m$ (identified by $mid$) and
$e$.

The mention $m$ identified by $mid$ should be correctly linked to one
and only one entity, $e$, given the sentence $s$.
Therefore, the similarity between the mention by $mid$ and its entity
$e$ being correctly linked, $\phi(mid, e)$, must be larger than the
similarity between the same mention and any other candidate
entities. To capture such correct linking, we design a margin-based
Hinge Loss function to distinguish the entity being correctly linked
from any other candidate entity, $e'$, that appears in $D_L$.
We give the loss function in Eq.~(\ref{elequ.lossf}).
\vspace*{-0.1cm}
\begin{equation} \label{elequ.lossf}
\setlength{\abovedisplayskip}{3pt}
\setlength{\belowdisplayskip}{3pt}
l = max \{0, 1 - [\phi(mid, e) - max_{e' \in V_{m} \land e' \neq e}
  \phi (mid, e')]\}
\end{equation}
where $V_{m}$ is the candidate entity set of mention $m$.
Here, Eq.~(\ref{elequ.lossf}) emphasizes the distinction between the
mapping entity and negative candidate entities.  It is important to
note that entities tend to be similar by the feature-mention embedding
(Eq.~(\ref{elequ.fecoh})) if they share more common features. In other
words, it is most likely that candidate entities for a mention achieve
certain degree of similarity, and it is hard to identify one correct
entity by using the feature-mention embedding. The mention-entity is
introduced to distinguish the mapping entity from other candidates.

Let $\textbf{m}$ and $\textbf{y}$ be vectors in $d$-dimensional
embedding space, for a mention $mid$ and an entity $e$, respectively.  We
use $l_{2}$-regularizations
to control the scale of the embeddings with which we represent the
Hinge Loss of mention-entity embedding by Eq.~(\ref{elequ.lossreg}).
\begin{equation} \label{elequ.lossreg}
\setlength{\abovedisplayskip}{3pt}
\setlength{\belowdisplayskip}{3pt}
MY = \sum l + \frac{\lambda}{2} \lVert \textbf{m} \rVert_{2}^{2} +
\frac{\lambda}{2} \lVert \textbf{y} \rVert_{2}^{2}
\end{equation}
where $l$ is the Hinge Loss (Eq.~(\ref{elequ.lossf})), and $\lambda$
is a parameter to control $l_2$-regularization.

\stitle{Knowledge Graph Embedding}: Let $G = (V, E)$ be a knowledge
graph that captures the relationships among entities. We make use of
$G$ to capture the relationships among entities, in a similar way like
the existing methods that model the coherence relations among entities
by $G$.
Let $e_{i}$ and $e_{j}$ be two entities in $G$.
In a similar way as to handle feature-entity embedding
(Eq.~(\ref{elequ.fecoh})), we use the skip-gram model for the
knowledge graph embedding, which is shown in Eq.~(\ref{elequ.eecoh}).
\vspace*{-0.3cm}
\begin{equation} \label{elequ.eecoh}
\setlength{\abovedisplayskip}{3pt}
\setlength{\belowdisplayskip}{3pt}
EE = - \sum_{e_{i} \in V}\sum_{e_{j} \in V} w_{ij} \cdot
\log p(e_{j} | e_{i})
\end{equation}
Also, we use the negative sampling to compute $\log p(e_{j} |
e_{i})$ as Eq.~(\ref{elequ.eens}) shows.
\vspace*{-0.3cm}
\begin{equation} \label{elequ.eens}
\setlength{\abovedisplayskip}{3pt}
\setlength{\belowdisplayskip}{3pt}
\log \sigma(\textbf{y}_{i}^{T}\textbf{y}'_{j}) + \sum_{k =
  1}^{Q}\Xi_{e_{j'}\thicksim P_{n}(e_{j})}[\log
  \sigma(-\textbf{y}_{i}^{T}\textbf{y}'_{j'})]
\end{equation}
Recall that two adjacency words in the Word2vec model act as the
context of each other. Here, $e_{i}$, $e_{j}$ can be regarded as the
context of each other as well. We represent an entity $e_{i}$ by two
vectors, $\textbf{y}_{i}$ and $\textbf{y}'_{i}$, where
$\textbf{y}_{i}$ is the target vector and $\textbf{y}'_{i}$ is the
context vector.
Note that $Q$ is the number of negative samples, and $\Xi$ is a
distribution.

To address the difference between mapping entity and other candidate
entities, we embed the ``is-a'' relation between entities and
types provided by the category hierarchy of a knowledge base.
%We use two ways to embed the relations. One is to
We use the skip-gram model, as given in Eq.~(\ref{elequ.etcoh}).  Then
we use negative sampling to compute $\log p(t | e)$ as given in
Eq.~(\ref{elequ.etns}).
\begin{equation} \label{elequ.etcoh}
\setlength{\abovedisplayskip}{3pt}
\setlength{\belowdisplayskip}{3pt}
ET = - \sum_{e \in V}\sum_{t \in V} w \cdot
\log p(t | e)
\end{equation}

\begin{equation} \label{elequ.etns}
\setlength{\abovedisplayskip}{3pt}
\setlength{\belowdisplayskip}{3pt}
\vspace*{-0.2cm}
\log \sigma(\textbf{y}^{T}\textbf{t}') + \sum_{k =
  1}^{Q}\Xi_{t'\thicksim P_{n}(t)}[\log
  \sigma(-\textbf{y}^{T}\textbf{t}')]
\end{equation}

\stitle{Coherence Embedding}: By coherence embedding, we consider two
mentions that appear in the same sentence, which are ignored in the
existing work.  Let $(m_i, e_i)$ and $(m_j, e_j)$ be two pairs of
(mention, entity) in the same sentence.  We observe that
such mention-entity pairs in the same sentence tend to be more
coherent with each other than those that appear in different sentences.
As an example, consider the two sentences S4 and S5 in
Figure~\ref{elfig:ekframe}(b).  ``Italian'' and ``German'' both occur
in S4, but ``Italian'' and ``Hamburg'' occur in S4 and S5
separately. ``Italian'' is considered to be more coherent to
``German'' than to ``Hamburg''. On the other hand, the entity
relations provided by a knowledge graph can address the relations between
mentions in different sentences. Take ``Italian'' in S4 and ``United
States'' in S5 as an instance. As shown in
Figure~\ref{elfig:ekframe}(c2), there is an edge connecting their
mapping entities ``Italy'' and ``United States'' in YAGO. Even though they
appear in different sentences, we know that they tend to be related
with each other. We represent these two insights between $(m_i, e_i)$
and $(m_j, e_j)$ using implication operation as
given in Eq.~(\ref{elequ.implication}).

\vspace*{-0.6cm}
\begin{equation} \label{elequ.implication}
\setlength{\abovedisplayskip}{3pt}
\setlength{\belowdisplayskip}{3pt}
(e_i, e_j) \rightarrow (m_{i}, m_{j})
\end{equation}
To embed
$f_{me} = (e_{i}, e_{j}) \to (m_{i}, m_{j})$,
by using the target vectors to represent $e_{i}$ and $e_{j}$, the
confidence of $f_{me}$ is given in Eq.~(\ref{elequ.conf}).
\begin{equation} \label{elequ.conf}
I(f_{me}) = I(e_{i}, e_{j})I(m_{i}, m_{j}) + 1 - I(e_{i}, e_{j})
\end{equation}
where $I(e_{i}, e_{j}) = \sigma ({\textbf{y}}_{i}^{T} \cdot
{\textbf{y}}_{j})$ and $I(m_{i}, m_{j}) = \sigma ({\textbf{m}}_{i}^{T}
\cdot {\textbf{m}}_{j})$.
We aim at maximizing $\sum_{f_{me} \in D_{L}}I(f_{me})$, and we do so
using the skip-gram model to compute the log-likelihood for
$I(f_{me})$ as given in Eq.~(\ref{elequ.if}).
\begin{equation} \label{elequ.if}
\setlength{\abovedisplayskip}{3pt}
\setlength{\belowdisplayskip}{3pt}
L(f_{me}) = - \log(I(f_{me})) - \sum_{k =
  1}^{Q}\Xi_{f'_{me}\thicksim P_{n}(w_{me})}[\log (1 -
  I(f'_{me}))]
\end{equation}
To find the negative sampling data $f'_{me}$, a quad $(e_{i}, e_{j'},
m_{i}, m_{j'})$ must satisfy two conditions: (1) $(m_{i}, m_{j'})$
do not exist in a sentence at the same time, and (2) there is no
edge between $e_{i}$ and $e_{j'}$ in the knowledge graph $G$.

\stitle{The Joint Embedding}: We unify the 5 embeddins discussed above
in a $d$-dimensional space. Thus, we formulate it as a joint
optimization problem showed in Eq.~(\ref{elequ.obj}).
\begin{equation} \label{elequ.obj}
\setlength{\abovedisplayskip}{3pt}
\setlength{\belowdisplayskip}{3pt}
\begin{aligned}
min \ O_{joint} = FE + MY + EE + ET \sum_{f_{me} \in D_{L}} L(f_{me}), \\
s.t.~{}\lVert \textbf{f} \rVert_{2} \le 1, \lVert \textbf{m} \rVert_{2} \le 1, \lVert \textbf{y} \rVert_{2} \le 1, \lVert \textbf{y}' \rVert \le 1, \lVert \textbf{t} \rVert \le 1
\end{aligned}
\end{equation}

\stitle{Model Learning}: To solve Eq.~(\ref{elequ.obj}), we adopt an
efficient stochastic sub-gradient descent algorithm
\cite{DBLP:journals/mp/Shalev-ShwartzSSC11} based on edge sampling
strategy \cite{DBLP:conf/www/TangQWZYM15}, which is also applied in
\cite{DBLP:conf/www/RenWHQVJAH17} for a joint embedding for type
inference. In each iteration, we alternatively sample from each of the
five objectives $\{FE, \sum l, EE, ET, \sum_{f_{me} \in D_{L}}
L(f_{me})\}$ a batch of edges (e.g., $(f, e)$) and their negative
samples, then update each embedding vector according to the
derivatives. Algorithm~\ref{elalg:ml} summarizes the model learning
process of our approach.

\stitle{Named Entity Linking}: With the learned embeddings by
Algorithm~\ref{elalg:ml}, we use any similarity measure suitable for
vectors to compute the similarity score between mentions and candidate
entities, as to map $\forall m \in M$ to $e \in V$.
First, we extract a context feature set $F_C(m)$ for $m$ in the same
way as we do for mentions in the training dataset $D_{L}$. And we use
a vector $\textbf{m}$ to represent $m$ using the learned embeddings of
$\textbf{f}_{m}$ in the form ${\textbf{m}} = \sum_{f_{m} \in F_{m}}
{\textbf{f}}_{m}$.
Second, we compute the dot product between ${\textbf{m}}$ and the
learned embedding vector of each candidante entity $e \in V_{m}$,
and select the candidate entity with the largest
value as the mapping entity.

%{\fontsize{10}{10}\selectfont
\begin{algorithm}[t]
\algsetup{linenosize=\tiny} \footnotesize
\caption{Joint Embedding on Named Entity Linking} \label{elalg:ml}
\textbf{Input:} generated training corpus $D_{L}$, knowledge graph
$G$, context features $F$, learning rate $\alpha$,
regularization parameter $\lambda$, number of negative samples $Q$,
dimension $d$\\
\textbf{Output:} mention embeddings $\{\textbf{m}\}$, entity target
and context embeddings $\{\textbf{y}\}$ and $\{\textbf{y'}\}$, context
feature embeddings $\{\textbf{f}\}$, type embeddings $\{\textbf{t}\}$

\begin{algorithmic}[1]
\STATE Initialize: vectors $\{\textbf{m}\}$, $\{\textbf{y}\}$,
$\{\textbf{y}'\}$, $\{\textbf{f}\}$, $\{\textbf{t}\}$ as random vectors;
\WHILE{it does not converge by Eq.~(\ref{elequ.obj})}
    \STATE Sample a feature-entity co-occurrence edge; select $Q$
    negative samples; update $\{\textbf{f}, \textbf{y}\}$ based on
    $FE$;
    \STATE Sample a mention $m$; get its mapping entity $e$; select
    $Q$ negative samples; update $\{\textbf{m}, \textbf{y}\}$ based on
    $l$;
    \STATE Sample an entity-entity edge; select $Q$ negative samples;
    update $\{\textbf{y}, \textbf{y}'\}$ based on $EE$;
    \STATE Sample an entity-type edge; select $Q$ negative samples;
    update $\{\textbf{y}, \textbf{t}\}$ based on $ET$;
    \STATE Sample a quad id; select $Q$ negative samples; update
    $\{\textbf{m}, \textbf{y}\}$ based on $L(f_{me})$
\ENDWHILE
\end{algorithmic}
\end{algorithm}
%}

\vspace*{-0.2cm}
\section{Experimental Evaluation} \label{elsec.exp}

% We call the proposed multi-source embedding approach for named entity
% linking as \textbf{Multi-Source Embedding}.

% \subsection{Dateset and Evaluation Metrics} \label{elsec.data}

In this experimental study, we use {\tt aida-yago2-dataset}
\cite{DBLP:conf/emnlp/HoffartYBFPSTTW11} to evaluate the performance of our method on sentence and document level
named entity linking.(1) The training dataset contains 946 documents. (2) the test
dataset contains 230 documents. The dataset contains about 20\%
unlinkable mentions those correct entities do not exist in YAGO2.

Following the existing methods, we ignore such unlinkable mentions and
only evaluate on the linkable mentions.

We use two metrics in the evaluation to evaluate the accuracy of a
proposed named entity linking method. One is Micro-averaging which
aggregates over all mentions in the dataset. The other is
Macro-averaging which aggregates on the input documents (or sentences) that a
document (or sentence) contains several mentions.

We evaluate our method at sentence level and document level.
To get the test sentences, we parse the testing documents into
sentences.  In the 230 testing documents in {\tt aida-yago2-dataset},
we get 2,380 sentences that contain at least one mention.

We compare our method with two state-of-art methods \textbf{AIDA}
\cite{DBLP:conf/emnlp/HoffartYBFPSTTW11} and \textbf{YSTT}
\cite{DBLP:conf/conll/YamadaS0T16}, as well as a strong baseline
\textbf{Prior Probability} which links a
mention to the entity with the largest prior probability from all
candidate entities.
For our approach, we set $\alpha$ = 0.02. $\lambda$ = 0.0001, $Q$ = 5
and $d$ = 300.

\vspace*{-0.2cm}
\subsection{Comparison with Other Methods}

\begin{table}
        \centering
        \caption{The Accuracy at Sentence and Document Level}
        \label{eltab.comp}
        \begin{tabular}{ccccc}
        \toprule
             & \multicolumn{2}{c}{Sentence Level} & \multicolumn{2}{c}{Document Level}\\
         Methods & \textbf{Micro} & \textbf{Macro} & \textbf{Micro} & \textbf{Macro}\\
         \midrule
         \textbf{Our approach} & 83.6\% & 82.3\% & 83.6\% & 84.1\% \\
         \textbf{YSTT} & 81.1\% & 79.6\% & 86.6\% & 87.1\% \\
    \textbf{AIDA} & 79.6\% & 78.4\% & 81.8\% & 81.9\% \\
    \textbf{Prior Probability} & 74.8\% & 76.7\% & 74.8\% & 77.9\% \\
    \bottomrule
        \end{tabular}
    \end{table}

Table~\ref{eltab.comp} shows the comparison results at sentence and document level.
Our method performs the best at sentence
level. It is due to the fact that we extract high quality context
features instead of using words. Although the context information is
very limited in a sentence, by learning from a large number of
training sentences (from Wikipedia), we can learn the relations
between the context features extracted from a sentence and the
candidate entities of a mention in the sentence. On document level, our method achieves
better performance than {\bf AIDA} and the baseline but worse than
{\bf YSTT}. This is because our method is designed for sentence level
named entity linking in order to address the issue that the same
mention in a document may map to different entities,
while {\bf YSTT} uses all the noun words in the whole
documents.

\vspace*{-0.3cm}
\subsection{Effect of Knowledge Base}

\begin{table}[t]%[!ht]
  \caption{Effect of Knowledge Base}
  \label{eltab.kb}
  \begin{tabular}{lll}
    \toprule
    \textbf{Measures} & \textbf{Micro} & \textbf{Macro} \\
    \midrule
    \textbf{Our approach} & 83.6\% & 82.3\% \\
    \textbf{Without $EE+ET$} & 80.1\% & 79.1\% \\
    \bottomrule
  \end{tabular}
\end{table}

We study the effect of knowledge base, which corresponds to the effect
of $EE+ET$ on the learned embeddings. Table~\ref{eltab.kb} shows the
effect of $EE+ET$ on the performance of our method.  From
Table~\ref{eltab.kb}, we can see that without the entity-entity
relations and entity-type relations provided by the knowledge base, the
learned embeddings become worse evaluated on the sentence level named
entity task. This is due to the lack of coherence among entities
across sentences.

\vspace*{-0.2cm}
\subsection{Entity Relatedness Performance}

To test the quality of learned embeddings of entities, we follow the
work \cite{DBLP:conf/cikm/CeccarelliLOPT13} to evaluate the
performance on the entity relatedness
task.
%The entity relatedness task is to find the relevant
%mention-entity pair from the candidate mention-entity pairs for a
%given mention-entity pair. We use dot product between two entity
%embeddings as their similarity score.
We compare our method with {\bf
  WLM} \cite{DBLP:conf/cikm/CeccarelliLOPT13} and entity embeddings
learned in {\bf YSTT} \cite{DBLP:conf/conll/YamadaS0T16} under three
standard metrics, which are NDCG@1, NDCG@5, NDCG@10
\cite{DBLP:journals/tois/JarvelinK02}, on the benchmark
dataset. Table~\ref{eltab.erp} shows the comparison results and our learned embeddings for entities
achieve the best performance on this task. This is because we use both
textual information and knowledge base which involve entities to
reinforce the learning of entity embeddings.

\begin{table}[t]%[!ht]
  \caption{Results on Entity Relatedness Task}
  \label{eltab.erp}
  \begin{tabular}{llll}
    \toprule
    \textbf{Measures} & \textbf{NDCG@1} & \textbf{NDCG@5} & \textbf{NDCG@10} \\
    \midrule
    \textbf{Our approach} & 68.0\% & 81.4\% & 82.0\% \\
    \textbf{YSTT} & 59\% & 56\% & 59\% \\
    \textbf{WLM} & 54\% & 52\% & 55\%\\
    \bottomrule
  \end{tabular}
\end{table}

\vspace*{-0.3cm}
\subsection{Case Study} We count the proportion that a document contains a mention which maps to more than one entities when it appears in different positions among the 1393 documents in {\tt aida-yago2-dataset}. We find that 127 documents in the whole dataset and 19 documents in test contain such mentions. Table~\ref{eltab.smde} shows the results on such mentions in the test data compared with {\bf YSTT} \cite{DBLP:conf/conll/YamadaS0T16} and Prior Probability on the document level. From Table~\ref{eltab.smde}, we can see our approach achieves the best performance. This indicates that our method is more powerful in distinguishing same mentions which map to different entities.

\begin{table}%[!ht]
  \caption{Results on Same Mention with Different Entities}
  \label{eltab.smde}
  \begin{tabular}{lll}
    \toprule
    \textbf{Measures} & \textbf{Micro} & \textbf{Macro} \\
    \midrule
    \textbf{Our approach} & 53.7\% & 49.6\% \\
    \textbf{YSTT} &46.3\% & 42.2\% \\
    \textbf{Prior Probability} & 35.8\% & 31.6\% \\
    \bottomrule
  \end{tabular}
\end{table}

\vspace*{-0.2cm}
\section{Conclusion}
\label{sec:conc}

In this work, we propose an unified embedding approach for named
entity linking by maximizing the relationships extracted from Wikipedia
and knowledge graph such as YAGO.
Our approach can link a mention in a sentence to a mapping entity with
highest accuracy. We donducted experimental studies using {\tt
  aida-yago2-dataset} which is also used in the state-of-art method
({\bf YSTT}) \cite{DBLP:conf/conll/YamadaS0T16}. Our approach
outperforms the state-of-art method in sentence level in our experimental studies.

\comment{
\section{Future Work}
There are two points to be improved in the future work.

First, the embedding of subject-predicate-object graph. So far, we do
not consider the influence of predicates on the learned embeddings of
entities. Next, we will design a projection method to connect the
embeddings of entities through the embeddings of predicates.

Second, on one hand, some sentences maybe contain only one mention. If
we only consider the sentence level coherence between (mention,
entity) pairs, a considerable mentions maybe have no positive
samples. On the other hand, if a sentence is too long, there may exist
some mentions in the sentence are unrelated with each other. Thus,
different levels of coherence between (mention, entity) pairs are
needed to be considered.
}

\vspace*{-0.2cm}
\bibliographystyle{ACM-Reference-Format}
\bibliography{sigproc}

\end{document}